\documentclass{article}

\usepackage[final]{nips_2016}

\usepackage[utf8]{inputenc} % allow utf-8 input
\usepackage[T1]{fontenc}    % use 8-bit T1 fonts
\usepackage{hyperref}       % hyperlinks
\usepackage{url}            % simple URL typesetting
\usepackage{booktabs}       % professional-quality tables
\usepackage{amsfonts}       % blackboard math symbols
\usepackage{nicefrac}       % compact symbols for 1/2, etc.
\usepackage{microtype}      % microtypography

\usepackage{algorithmic}
\usepackage{amstext}
\usepackage{amsmath}
\usepackage{graphicx}
\usepackage[export]{adjustbox}
\usepackage{subcaption}
\usepackage{float}
\newcommand{\dir}{\text{Dirichlet}}
\newcommand{\mult}{\text{Multinomial}}

\newcommand{\norm}[1]{\left\lVert#1\right\rVert}

\title{A Nonparametric Latent Factor Model For Location-Aware Video Recommendations}

\author{
  Ehtsham Elahi \\
  Algorithms Engineering\\
  Netflix, Inc.\\
  Los Gatos, CA 95032 \\
  \texttt{eelahi@netflix.com} \\
}

\begin{document}

\maketitle

\begin{abstract}
We are interested in learning customers' video preferences from their historic viewing patterns and geographical location. We consider a Bayesian latent factor modeling approach for this task. In order to tune the complexity of the model to best represent the data, we make use of Bayesian nonparameteric techniques. We describe an inference technique that can scale to large real-world data sets. Finally we show results obtained by applying the model to a large internal Netflix data set, that illustrates that the model was able to capture interesting relationships between viewing patterns and geographical location.
\end{abstract}

\section{Introduction}
In a web application we are provided with a rich view of each user. For example in a video streaming application, like Netflix, we can observe not only their preference for different types of content but also how those preferences change with respect context, such as time of day, day of week, device, and so on. An important contextual variable that influences a customer's preferences is their geographical location. It is reasonable to assume that customers who live in close proximity may have similar viewing preferences. Hence, a model is required that can capture not only a customer's latent viewing preferences, but also the relationship between those and their location. To capture both these aspects we seek to model them in a unified model so that both location and viewing behavior can take advantage of information in each modality. For this task, we employ a nonparametric latent factor model to jointly model a customer's viewing history and their geographical location.

Nonparametric mixed membership style techniques have shown great promise in modeling large collections of documents [1]. Given that there is more information available for a document (author, date of publishing, metadata etc.) than just its content, it seems natural to extend these approaches to model all these modalities in a unified approach. Hence there have been many attempts in applying nonparametric latent factor modeling for such data sets [2]. Our approach uses a similar model structure as [2] which attempts to model document-level features along with the content of documents. For the problem under consideration, we view a customer's viewing history as an unordered collection of discrete view events from Netflix's video catalog. Geographical locations of customers are expressed in longitudes and latitudes. The geographical locations can be viewed as points on a 2-sphere. Therefore we use an approach similar to [3] using Von Mises-Fisher distribution to describe geographical data. The full model combines these sub-components (viewing history and geographical location) and is able to learn embeddings for customers' viewing history data, geographical location data, and the interactions between the two.

The following sections detail how we model these components, how we infer that model (in a way that scales to large-scale data sets), and finally results from an internal Netflix data set that illustrates that the model is indeed able to capture interactions between geographical information and viewing preferences.

\section{Model Details}
\label{model_details}
The component of our model which describes customers' streaming history data is a nonparametric mixed membership model that uses a hierarchical dirichlet process to learn latent video factors; each of which are multinomial distributions over content catalog. Similarly, the component of our model that models geographical locations uses hierarchical dirichlet process to learn latent factors for geographical locations; each of which are Von Mises-Fisher distributions over a 2-sphere. Finally the relationship between the two latent spaces is expressed through a dirichlet process over the interaction of video and location latent factor spaces. We summarize our modeling assumptions as follows and then comment on different components of the model.
\\
\begin{algorithmic}[1]
\STATE $\phi_0 \sim \text{DP}(. |\ \alpha_{\phi_0}, H(\mu, c)), \pi_0 \sim \text{DP}(. |\  \alpha_{\pi_0}, H(\beta))$
\STATE $\omega \sim \text{DP}(. | \alpha_{\omega}, H(\text{DP}( . | \alpha_{\phi}, \phi_0)\text{ x DP}( . | \alpha_{\pi}, \pi_0)))$
\FOR{customer $d$ in data set $D$}
  \STATE $(\phi_d, \pi_d) \sim~ \omega$
  \STATE $\mu_d, c_d \sim \phi_d$
  \STATE $\text{location}_d\ \sim \text{Von Mises-Fisher}(. | \mu_{d}, c_{d})$ 
  \FOR{$j_d$ in video history $J_d$}
  \STATE $\beta_{j_d}\ \sim \pi_d$
  \STATE $v_{j_d}\ \sim \mult( . |\ \beta_{j_d}, 1) $.
  	\ENDFOR
 \ENDFOR
\end{algorithmic}

\subsection{Modeling Location Data}
For geographical location data of customers, we need a distribution which can express the spherical nature of the data. We make use of Von Mises-Fisher distribution for modeling locations. We use the following parameterization of Von Mises-Fisher distribution:
\begin{align}
\Pr(x | \mu, c) &= C_{D}(c) \exp(c\mu^Tx)
\end{align}
where $C_D(c) =\frac{c^{0.5D - 1}}{(2\pi)^{0.5D} I_{0.5D - 1}(c)}$; $\mu$ and $c$ are the parameters of the distributions; $I_{0.5D - 1}(c)$ is modified Bessel function of first kind with order $0.5D - 1$ computed at $c$. This parameterization requires locations to be expressed in Euclidean coordinates. Hence, we convert geo-spherical coordinates to Euclidean system. The prior distributions for $\mu$ and $c$ are:
\begin{align}
\Pr(\mu | \mu_0, c_0) &= \text{Von Mises-Fisher}(\mu | \mu_0, c_0)\\
\Pr(c | m_c, \sigma_c) &= \text{logNormal}(c | m_c, \sigma_c)
\end{align}
The prior distribution of $\mu$ is chosen to be a Von Mises-Fisher Distribution itself which is conjugate to Von Mises-Fisher likelihood. The concentration parameter $c$ does not have a conjugate prior. We use a log normal prior for $c$ similar to [3].
\subsection{Modeling Video History Data}
As mentioned above, we view customers' videos streaming history as unordered collections of videos watched from the Netflix's catalog. We use a Dirichlet-Multinomial conjugate model for representing video streaming history of customers:
\begin{align}
\Pr(v | \beta) &= \mult(v | \beta, 1)\\
\Pr(\beta | \gamma) &= \dir(\beta | \gamma)
\end{align}
$\mult(v | \beta, 1)$ represents a single draw from the multinomial distribution on a video catalog of size \textbf{V}.
\subsection{Modeling Interaction of Video and Location Latent Factors}
The interaction of video and geographical latent spaces is modeled by a dirichlet process with a product base measure $\text{DP}(. | \alpha_{\phi}, \phi_0)\ \text{x}\ \text{DP}(. | \alpha_{\pi}, \pi_0)$ i-e the base measure is on atoms which are pairs of dirichlet processes drawn from the dirichlet process on location and video latent factors respectively. This construction allows the model to flexibly learn as many interactions between video preferences and geo-locations as needed to best express the data. 

\subsection{Inference}
We use a sampling based approach for posterior inference. Due to dirichlet-multinomial conjugacy in the video component of the model, we  collapse out $\beta$ for each latent video factor. For the location component of the model, prior distribution of $\mu$ (Von Mises-Fisher) is conjugate to Von Mises-Fisher likelihood, hence we collapse out $\mu$ as well for each latent location factor. The prior distribution of $c$ (log-normal) is not conjuage to Von Mises-Fisher likelihood, hence we use Metropolis-Hasting algorithm to sample $c$ for each latent location factor. For the nonparametric components, we make use of the direct assignment scheme described in [1]. Hence, instead of sampling atoms, we sample indicators to those atoms. Specifically, $t_d$ (taking values in t = 1,...,$\infty$) is the indicator to the atom $(\phi_{t_d}, \pi_{t_d})$, $s_d$ (taking values in s = 1,...,$\infty$) is the indicator to the atom $(\mu_{s_d}, c_{s_d})$, and $z_{j_d}$ (taking values in z = 1,...,$\infty$) is the indicator to the atom $\beta_{z_{j_d}}$. Additionally, we sample the global dirichlet processes $\phi_0$ and $\pi_0$ according to the direct assignment scheme in [1]. The sampling distributions for these latent variables are as follow:

\begin{align}
\Pr(t_d = t\ |\ ...) &\propto (n_t^{-d})
\left(\frac{n_{t,s_d}^{-d} + \alpha_{\phi}\phi_{0,s_d}}{n_{1_{t,.}}^{-d} + \alpha_{\phi}}\right)
\prod_{j_d = 1}^{J_d}\left(\frac{n_{t,z_{j_d}}^{-d} + \alpha_{\pi}\pi_{0,z_{j_d}}}{n_{2_{t,.}}^{-d}+\alpha_{\pi}}\right)\\
\Pr(t_d = t^{\text{new}}\ |\ ...) &\propto (\alpha_{\omega})(\phi_{0,s_d})\prod_{j_d=1}^{J_d}(\pi_{0,z_{j_d}})\\
\Pr(s_d = s\ |\ ...) &\propto (n_{t_d,s}^{-d} + \alpha_{\phi}\phi_{0,s})
C_{D}(c_s)\left(\frac{C_{D}\norm{c_s\sum_{l : l \neq d, s_l  = s} \text{location}_l + c_0\mu_0}}{C_{D}\norm{c_s\sum_{l : s_l = s} \text{location}_l + c_0\mu_0}}\right)\\
\Pr(s_d = s^{\text{new}}\ |\ ...) &\propto (\alpha_{\phi}\phi_{0,s^{\text{new}}})
C_{D}(c_{s^\text{new}})\frac{C_{D}\norm{c_0\mu_0}}{C_{D}\norm{c_{s^\text{new}}\text{location}_d + c_0\mu_0}}\\
\Pr(z_{j_d} = z\ | ...) &\propto (n_{t_d,z}^{-j_d} + \alpha_{\pi}\pi_{0,z})
\left(\frac{n_{z,v_{j_d}}^{-v_{j_d}} + \gamma_{v_{j_d}}}{n_{z,.}^{-v_{j_d}} + \sum_{v = 1}^{V}\gamma_{v}}\right)\\
\Pr(z_{j_d} = z^{\text{new}}\ | ...) &\propto (\alpha_{\pi}\pi_{0,z^{\text{new}}})
\left(\frac{\gamma_{v_{j_d}}}{\sum_{v = 1}^{V}\gamma_{v}}\right)\\
\Pr(c_s\ | ...) &\propto \text{log-normal}(c_s | m_c, \sigma_c)
\frac{(C_D(c_s))^{n_s}C_D(c_0)}
{C_D(\norm{c_s\sum_{d : s_d = s}\text{location}_d + c_0 \mu_0})}
\end{align}
Above, $\Pr(\text{variable} | ...)$ represents the complete conditional distribution of the variable. Notations like $n_{t,s_d}^{-d}$ represent conditional counts; count of variables $t$ and $s_d$ ignoring customer $d$ for example. Notations like $n_{1_{t,.}}^{-d}$ and $n_{2_{t,.}}^{-d}$ represent marginal counts; marginal counts of variable $t$, marginalizing over $s_d$ and $z_{j_d}$ respectively for all customers except $d$ (subscripts $1$ and $2$ are used to differentiate the two marginals involving $t$).

\section{Experiments}
In order to scale our sampling based posterior inference, we use an approximate parallel gibbs sampling approach as described in [4]. For our experiment we use an internal data set that contains video viewing history for one million Netflix customers along with their geographical locations. We include some of the examples of latent video and geographical factor learned by our model as well as the top three video topics for the two geographical latent factors found in the United States of America.
\begin{figure}[h]
  \begin{subfigure}{0.5\textwidth}
  \includegraphics[scale = 0.19, left]{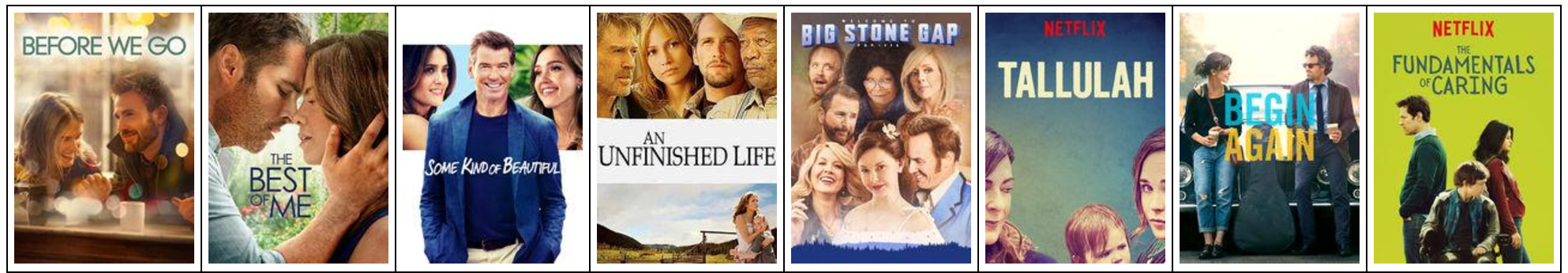}
  \caption{Romantic Shows Topic}
  \end{subfigure}
  \begin{subfigure}{0.5\textwidth}
  \includegraphics[scale = 0.19, right]{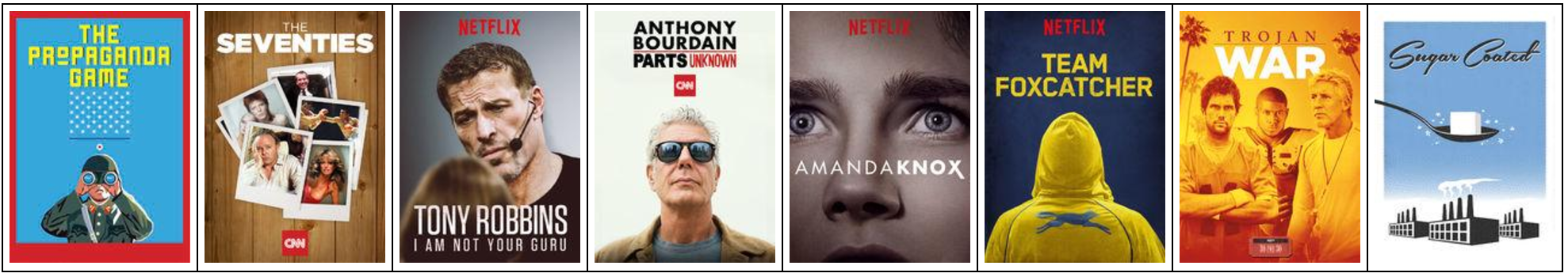}
  \caption{Documentaries Topic}
  \end{subfigure}
  
  \caption{Video Latent Factors capturing Romantic Shows and Documentaries}
\end{figure}

\begin{figure}[H]
  \centering
  \includegraphics[scale=.35]{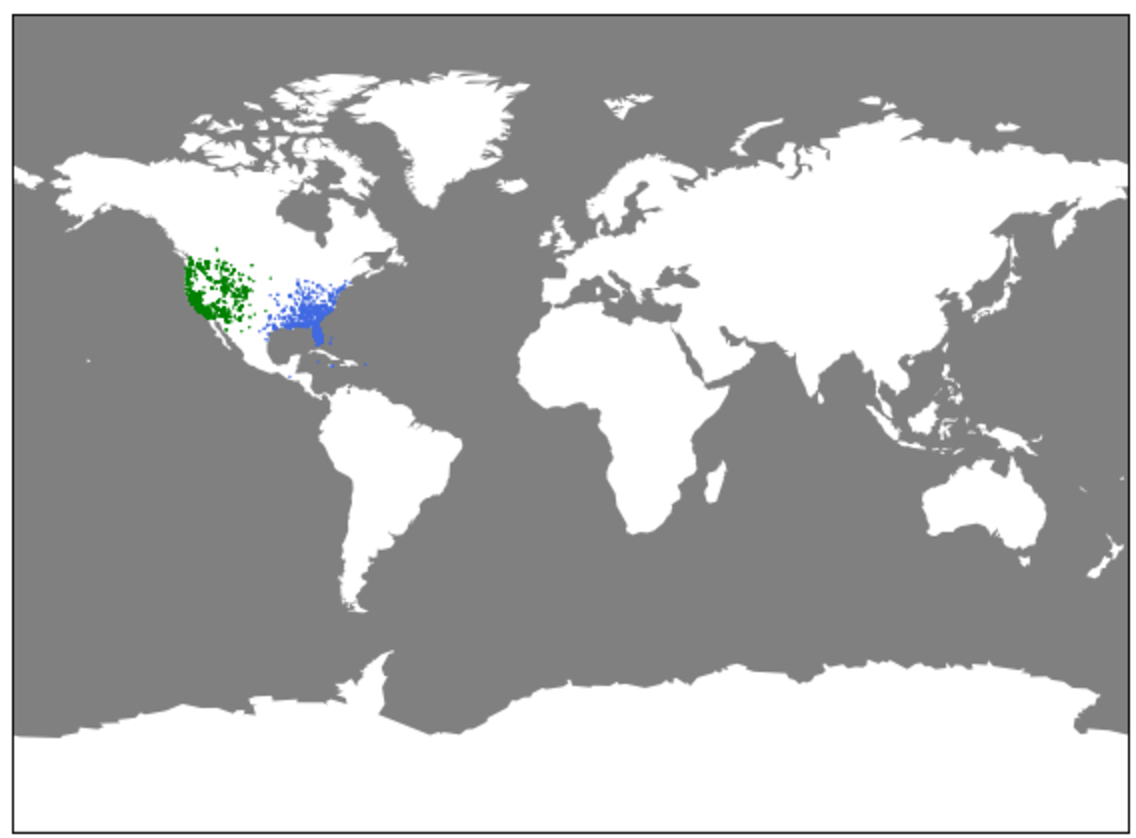}
  \caption{Two Example Geographical Latent Factors Found in the United States of America.}
\end{figure}

\begin{figure}[H]
  \begin{subfigure}{0.5\textwidth}
  \includegraphics[scale=0.8, left]{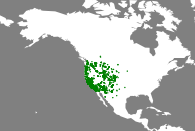}
  \end{subfigure}
  \begin{subfigure}{0.5\textwidth}
  \includegraphics[scale=0.19, right]{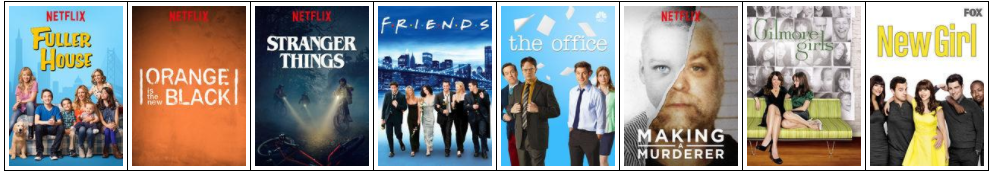}
  \includegraphics[scale=0.19, right]{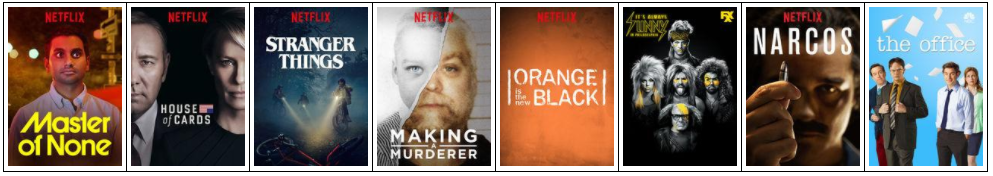}
  \includegraphics[scale=0.19, right]{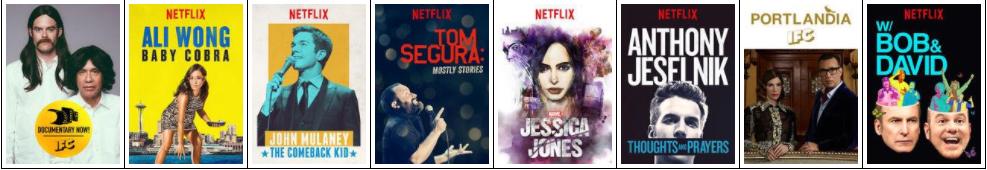}
  \end{subfigure}
  \caption{Top Video Topics for the geographical latent factor}
\end{figure}

\begin{figure}[H]
  \begin{subfigure}{0.5\textwidth}
  \includegraphics[scale=0.8, left]{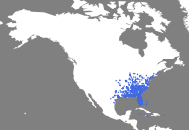}
  \end{subfigure}
  \begin{subfigure}{0.5\textwidth}
  \includegraphics[scale=0.19, right]{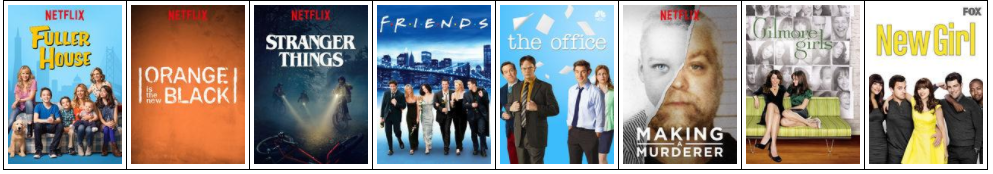}
  \includegraphics[scale=0.19, right]{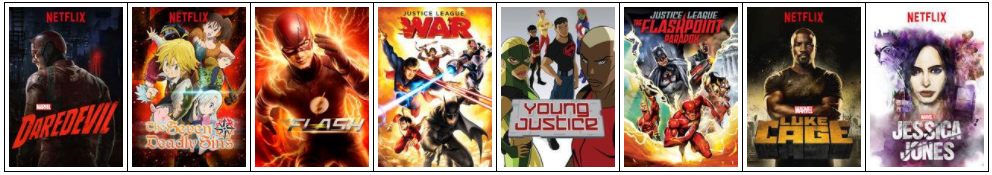}
  \includegraphics[scale=0.19, right]{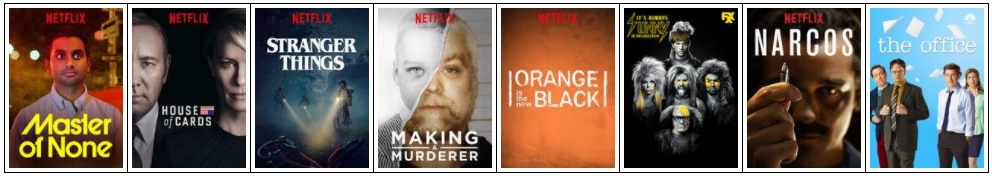}
  \end{subfigure}
  \caption{Top Video Topics for the geographical latent factor}
\end{figure}

\section{Conclusion}
We use bayesian non-parameteric machinery to combine geographical and viewing behavior information of customers of Netflix for location-aware video recommendations. The approach presented can also be helpful in situations where the viewing history data is sparse or cold-start scenario.
\section*{References}
\medskip
\small

[1] Teh, Y.W., Jordan, M.I.,Beal, M.J.,\ \& Blei, D.M.\ (2006) Hierarchical Dirichlet Process. {\it Journal of the American Statistical Association,} 101, 1566-1581.

[2] Nguyen, V., Phung, D., Nguyen, X.,Venkatesh, S.\ \& Bui, H.H.\ (2014) Bayesian Nonparametric Multilevel clustering with group-level contexts {\it . Proceedings of the ICML} 

[3] Gopal, S.\ \& Yang, Y.\ (2014) Von Mises-Fisher Clustering Models. {\it Proceedings of the ICML}.

[4] Newman, D., Asuncion, A., Smyth, P.\ \& Welling, M.\ (2009) Distributed Algorithms for Topic Models. {\it Journal of Machine Learning},{\bf 10}(Aug):1801-1828.

\end{document}